\documentclass{ecai} 
\usepackage{latexsym}
\usepackage{amssymb}
\usepackage{amsmath}
\usepackage{amsthm}
\usepackage{booktabs}
\usepackage{enumitem}
\usepackage{graphicx}
\usepackage{color}
\usepackage{float}
\usepackage{caption}
\usepackage[none]{hyphenat}  
\usepackage{microtype}       
\usepackage[numbers]{natbib}

\newcommand{\BibTeX}{B\kern-.05em{\sc i\kern-.025em b}\kern-.08em\TeX}

\begin{document}

%%%%%%%%%%%%%%%%%%%%%%%%%%%%%%%%%%%%%%%%%%%%%%%%%%%%%%%%%%%%%%%%%%%%%%%%

\begin{frontmatter}

%%% Use this command to specify your submission number.
%%% In doubleblind mode, it will be printed on the first page.

\paperid{123} 

%%% Use this command to specify the title of your paper.

\title{Single Domain Generalization in Diabetic Retinopathy: A Neuro-Symbolic Learning Approach  }

%%% Use this combinations of commands to specify all authors of your 
%%% paper. Use \fnms{} and \snm{} to indicate everyone's first names 
%%% and surname. This will help the publisher with indexing the 
%%% proceedings. Please use a reasonable approximation in case your 
%%% name does not neatly split into "first names" and "surname".
%%% Specifying your ORCID digital identifier is optional. 
%%% Use the \thanks{} command to indicate one or more corresponding 
%%% authors and their email address(es). If so desired, you can specify
%%% author contributions using the \footnote{} command
%%% Use this environment to include an abstract of your paper.

\author{
\fnms{Midhat}~\snm{Urooj},
\fnms{Ayan}~\snm{Banerjee},
\fnms{Farhat}~\snm{Shaikh},
\fnms{Kuntal}~\snm{Thakur},
\fnms{Sandeep}~\snm{Gupta}
}

\address{
Impact Lab, Arizona State University, Tempe, AZ, USA. \\
Emails: \texttt{\{murooj, abanerj3, fshaik12, kthakur9, sandeep.gupta\}@asu.edu}
}

\begin{abstract}
Domain generalization remains a critical challenge in medical imaging, where models trained on single sources often fail under real-world distribution shifts. We propose  \textbf{KG-DG}, a neuro-symbolic framework for diabetic retinopathy (DR) classification that integrates vision transformers with expert-guided symbolic reasoning to enable robust generalization across unseen domains. Our approach leverages clinical lesion ontologies through structured, rule-based features and retinal vessel segmentation, fusing them with deep visual representations via a confidence-weighted integration strategy. The framework addresses both single-domain generalization (SDG) and multi-domain generalization (MDG) by minimizing the KL divergence between domain embeddings, thereby enforcing alignment of high-level clinical semantics. 

Extensive experiments across four public datasets (APTOS, EyePACS, Messidor-1, Messidor-2) demonstrate significant improvements: up to a 5.2\% accuracy gain in cross-domain settings and a 6\% improvement over baseline ViT models. Notably, our symbolic-only model achieves a 63.67\% average accuracy in MDG, while the complete neuro-symbolic integration achieves the highest accuracy compared to existing published baselines and benchmarks in challenging SDG scenarios. Ablation studies reveal that lesion-based features (84.65\% accuracy) substantially outperform purely neural approaches, confirming that symbolic components act as effective regularizers beyond merely enhancing interpretability. Our findings establish neuro-symbolic integration as a promising paradigm for building clinically robust, and domain-invariant medical AI systems.
\end{abstract}
\end{frontmatter}

%%%%%%%%%%%%%%%%%%%%%%%%%%%%%%%%%%%%%%%%%%%%%%%%%%%%%%%%%%%%%%%%%%%%%%%%
\section{Introduction}
Diabetic Retinopathy (DR) is a microvascular complication of Diabetes Mellitus that affects the retinal vasculature, leading to hemorrhages, microaneurysms, exudates, and cotton-wool spots which, if left untreated, can culminate in irreversible vision loss \cite{khandelwal2023}. Manual grading of fundus photographs by expert ophthalmologists remains the clinical gold standard but is both time-consuming and subject to inter-observer variability \cite{kauppi2019}. Despite the success of deep learning models—particularly Vision Transformers (ViTs)—on single-source DR datasets \cite{dosovitskiy2021,zhao2022}, their performance suffers when confronted with domain shifts caused by variations in imaging devices, resolution settings, and patient demographics. Although Domain Generalization (DG) strategies such as Empirical Risk Minimization under the DomainBed protocol \cite{gulrajani2021} offer a baseline for robustness, they often overlook the integration of structured clinical knowledge and realistic augmentation techniques that are critical for reliable cross-domain deployment.

Neuro-symbolic learning, which integrates deep learning with symbolic reasoning, has gained traction as a promising strategy to improve domain generalization in medical imaging. Deep models extract complex patterns from raw data, while symbolic components encode high-level domain knowledge and constraints, thereby effectively guiding model behavior across varying domains. This hybrid approach can mitigate overfitting to domain-specific artifacts by enforcing consistency with known anatomical or pathological rules. For example, Han et al. introduced a neuro-symbolic framework for spinal MRI segmentation that embeds anatomical priors into a deep adversarial graph network, resulting in better generalization and interpretability across different datasets \cite{han2021}. Similarly, Ozkan and Boix demonstrated that training across multiple imaging modalities (e.g., MRI, CT, ultrasound) significantly improves generalization to unseen domains, emphasizing the value of diverse training data and domain-aware learning strategies \cite{ozkan2024}. These findings suggest that symbolic reasoning components can serve as a regularizing force that biases models toward clinically meaningful and domain-invariant features—thereby enabling more robust, scalable medical AI systems.

Despite recent advances, most current neuro-symbolic methods remain narrowly focused—typically emphasizing symbolic reasoning mechanics without incorporating the type of clinical knowledge used by medical experts to inform robust, generalizable decision-making. In particular, few approaches simultaneously address both symbolic knowledge integration and domain generalization in a cohesive framework. This gap motivates our proposed method, \textbf{KG-DG}, a knowledge-guided domain generalization framework that unifies structured clinical knowledge with deep learning models in a scalable  manner. KG-DG encodes domain-invariant biomarkers—such as exudates, hemorrhages, and vascular abnormalities—directly into the learning pipeline, guiding classification tasks while enhancing out-of-distribution (OOD) robustness. Unlike prior approaches, our framework generalizes across imaging modalities, as demonstrated by strong performance in both diabetic retinopathy classification and seizure-onset detection from MRI scans. By embedding clinical expertise at the architectural level, KG-DG bridges the gap between symbolic interpretability and neural robustness, advancing the development of domain-generalizable medical AI.

%%%%%%%%%%%%%%%%%%%%%%%%%%%%%%%%%%%%%%%%%%%%%%%%%%%%%%%%%%%%%%%%%%%%%%%%
\section{Related Work}

Vision transformers (ViTs) have revolutionized medical image analysis, particularly in ophthalmology, offering a powerful alternative to traditional convolutional neural networks. Dosovitskiy et al. \cite{dosovitskiy2021} established the foundation by demonstrating ViTs' state-of-the-art performance on large-scale image recognition benchmarks, catalyzing their adoption for diabetic retinopathy (DR) detection. Subsequent work by Kothari et al. introduced TransDR \cite{wang2024diffusion}, enhancing ViTs with lesion-aware attention mechanisms that improve lesion localization capabilities, though without explicitly addressing domain shift robustness challenges.

The challenge of domain generalization (DG) has become increasingly prominent in medical AI deployment. DomainBed benchmarks \cite{gulrajani2021} have illuminated the fundamental difficulty of ensuring that deep models trained on source domains maintain reliable performance on unseen target domains. Gulrajani and Lopez-Paz's findings revealed that even sophisticated methods often fail to outperform simple empirical risk minimization (ERM) under rigorous evaluation protocols, emphasizing the critical importance of careful DG methodology design.

The integration of structured clinical knowledge represents a promising direction for enhancing both interpretability and performance in medical imaging systems. GraphDR, developed by Khandelwal et al. \cite{khandelwal2023}, exemplifies this approach by leveraging diabetic retinopathy lesion ontologies within graph convolutional networks to guide feature learning processes. Similar knowledge-driven strategies have been successfully applied to chest X-ray and histopathology classification tasks, grounding models in established disease relationships. These developments suggest that embedding lesion ontologies into transformer self-attention mechanisms could provide an effective pathway for learning clinically meaningful and domain-invariant representations.

Contrastive learning methodologies have emerged as another approach for addressing domain generalization challenges by disentangling domain-specific artifacts from semantic features. Deep CORAL \cite{sun2016} pioneered domain alignment through feature covariance alignment in convolutional neural networks, while contemporary DG methods employ cross-domain contrastive losses to promote separation between invariant and domain-dependent representations. Chen et al. \cite{chen2024} successfully extended these principles to medical imaging, achieving improved generalization in histopathology classification under distributional shifts. Nevertheless, contrastive disentanglement approaches remain relatively unexplored in diabetic retinopathy classification tasks.

Our approach bears conceptual similarity to Concept Bottleneck Models \cite{koh2020} and their quantitative extensions, such as Concept Embedding Models \cite{espinosa2022}. While these models learn to predict labels through intermediate human-defined concepts, our KG-DG framework differs by explicitly combining clinical rules with learned neural embeddings via confidence fusion, rather than strictly bottlenecking predictions through concepts. Compared to GraphDR \cite{khandelwal2023}, which employs lesion ontologies in a Graph Convolutional Network (GCN), our work emphasizes hybrid neuro-symbolic fusion at the decision level, making it more adaptable to diverse backbones and datasets. 
%%%%%%%%%%%%%%%%%%%%%%%%%%%%%%%%%%%%%%%%%%%%%%%%%%%%%%%%%%%%%%%%%%%%%%%%

\subsection{Domain Generalization}
In many real-world applications, particularly in biomedical fields, it is unrealistic to expect access to new patients' data before model deployment due to domain shifts between data from different patients \cite{muandet2013}. To address this challenge, the concept of Domain Generalization (DG) was introduced \cite{blanchard2011}. DG aims to train models on data from one or more related but distinct source domains, enabling them to generalize effectively to unseen, out-of-distribution (OOD) target domains. Since its formal introduction by Blanchard et al. in 2011 \cite{blanchard2011}, a wide range of techniques have been proposed to tackle the DG challenge \cite{zhou2021mixstyle}--\cite{SWAD}.

These approaches include learning domain-invariant representations by aligning source domain distributions \cite{li2018domain,li2018deep}, simulating domain shifts during training using meta-learning \cite{li2018learning,balaji2018metareg}, and generating synthetic data through domain augmentation \cite{zhou2020learning,zhou2020deep}. From an application perspective, DG has been explored in various areas such as computer vision (e.g., object recognition \cite{li2017deeper,li2019feature}, semantic segmentation \cite{volpi2019addressing}, and person re-identification \cite{zhou2021mixstyle,zhou2020learning}), speech recognition \cite{shankar2018generalizing}, natural language processing \cite{balaji2018metareg}, medical imaging \cite{liu2020msnet,liu2020shape}, and reinforcement learning \cite{zhou2021mixstyle}.

Unlike related paradigms like domain adaptation or transfer learning, DG uniquely addresses scenarios where no target domain data is accessible during training. The original motivation behind DG stemmed from a medical application known as automatic gating in flow cytometry \cite{blanchard2011}. This technique involves classifying cells in blood samples—such as distinguishing lymphocytes from non-lymphocytes—based on measured cellular properties. While such automation can significantly aid in diagnosis by replacing the labor-intensive and expert-dependent manual gating process, patient-specific distribution shifts hinder model generalization. Collecting new labeled data for every patient is not feasible, thereby underscoring the need for DG solutions.

In medical imaging, domain shift is especially prevalent due to variations across clinical sites and individual patients \cite{liu2020shape,dou2019domain}. Datasets like Multi-site Prostate MRI Segmentation \cite{liu2020shape} and Chest X-rays \cite{mahajan2021domain} reflect this reality, with differences in imaging equipment and acquisition protocols introducing substantial distribution variability.

Domain alignment techniques have been applied across diverse DG tasks, including object recognition \cite{li2018deep,ghifary2015domain}, action recognition \cite{li2018domain}, face anti-spoofing \cite{shao2019multi,jia2020single}, and medical image analysis \cite{li2020domain,aslani2020scanner}. Among the simplest and most effective strategies for mitigating domain shift in medical imaging are image transformations \cite{otalora2019staining,chen2020improving,zhang2020generalizing}. These transformations can emulate changes in color and geometry, often caused by device heterogeneity—such as different scanners in various medical centers. However, care must be taken in selecting these transformations. In some domains (e.g., digit recognition or optical character recognition), certain transformations like horizontal or vertical flips may alter the semantic label, leading to label shift. Therefore, transformation-based strategies must be chosen judiciously to preserve task relevance and integrity.

%%%%%%%%%%%%%%%%%%%%%%%%%%%%%%%%%%%%%%%%%%%%%%%%%%%%%%%%%%%%%%%%%%%%%%%%

%%%%%%%%%%%%%%%%%%%%%%%%%%%%%%%%%%%%%%%%%%%%%%%%%%%%%%%%%%%%%%%%%%%%%%

\begin{table}[t]
\centering
\caption{Clinical Signs of DR and Their Diagnostic Significance}
\label{tab:dr_symptoms}
\resizebox{\columnwidth}{!}{%
\begin{tabular}{lp{0.58\columnwidth}}
\toprule
\textbf{Symptom} & \textbf{Key Observations and Diagnostic Relevance} \\
\midrule
Microaneurysms & 
Tiny red capillary dilations in the retina; the earliest sign of Mild NPDR. Their progression correlates with disease severity \citep{frank2004, wilkinson2003, singh2008}. \\

Haemorrhages & 
Includes dot/blot and flame-shaped types indicating microvascular leakage. Severe NPDR is marked by $>$20 hemorrhages in all quadrants; risk of PDR rises to ~50\% within a year \citep{aao2023, statpearls2024, etdrs1991, singh2008}. \\

Hard Exudates & 
Lipid-rich deposits from chronic leakage, often in/near the macula. Indicative of risk for Diabetic Macular Edema (DME), a major cause of vision loss \citep{etdrs1991, statpearls2024, shukla2025}. \\

Cotton Wool Spots & 
Fluffy white retinal lesions caused by nerve fiber layer infarctions. Signify retinal ischemia in Moderate to Severe NPDR \citep{frank2004, statpearls2024, shukla2025}.\\

Subhyaloid Haemorrhages & 
Boat- or D-shaped hemorrhages between retina and hyaloid face, typically from ruptured neovascular vessels. Hallmark of Proliferative DR \citep{yanoff2019, aapos2023, shukla2025}. \\

Neovascularization & 
Fragile vessel growth on optic disc (NVD) or retina (NVE). Defining trait of PDR. High-risk cases without treatment face ~50\% vision loss within 5 years \citep{etdrs1991, aao2023, shukla2025}.
\\
\bottomrule
\end{tabular}
}
\end{table}

\begin{figure}[ht]
    \centering
    \includegraphics[width=\linewidth]{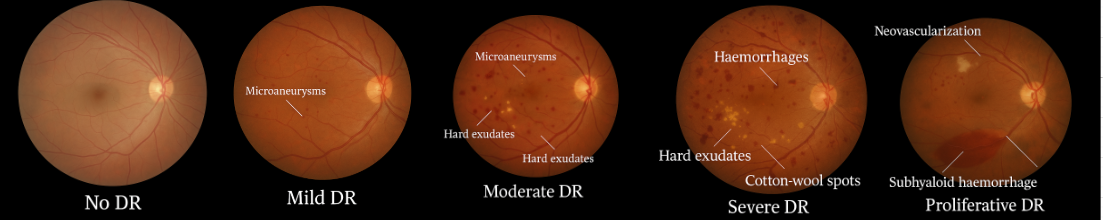}
    \captionsetup{justification=raggedright, singlelinecheck=false, skip=2pt}
    \caption{Fundus images showing Diabetic Retinopathy progression: from No DR to Proliferative DR, highlighting key lesions at each stage  \cite{kauppi2019}.}
    \label{fig:dr_stages}
    \vspace{\baselineskip}
\end{figure}
\subsection{The Role of Knowledge-Guided Systems in Improving Model Generalization}

Deep learning models, although powerful in capturing visual patterns, are notoriously sensitive to domain shifts—arising from demographic, device, or protocol variability—especially in clinical imaging tasks. Knowledge-guided systems mitigate this by embedding expert-defined symbolic rules, lesion ontologies, and diagnostic priors into the learning process. These structured features encode semantically meaningful, domain-stable concepts such as "presence of neovascularization implies Proliferative DR," thereby reducing reliance on superficial correlations.

Prior works substantiate this claim. GraphDR \cite{khandelwal2023} incorporated a lesion graph into a GCN to improve generalization on retinal datasets. Similarly, DeepXSOZ \cite{kamboj2023deepxsoz} utilized symbolic EEG biomarkers, achieving a 38.5\% performance gain over CNN baselines for seizure onset prediction. Symbolic abstraction enhances out-of-distribution robustness, as clinical definitions (e.g., "cotton-wool spots") are invariant across hospitals even when imaging characteristics differ. These features are extracted via interpretable modules like YOLOv11 and retinal vein segmentation networks, which we employ in our framework as open-source, plug-and-play components.

Beyond robustness, knowledge guidance facilitates meaningful domain alignment. In KG-DG, Kullback-Leibler (KL) divergence between domain embeddings—constructed from symbolic features—is minimized, improving convergence by aligning high-level clinical semantics. This complements domain generalization strategies like IRM \cite{arjovsky2019invariant} and Fishr \cite{rame2022fishr}, which rely on invariant predictors across environments.

%%-------------------------------------------------------
\section{Methodology}

We propose a general-purpose framework for \textit{knowledge imputation into AI-based models}, enabling integration of clinically validated rules, visual biomarkers, and demographic insights into conventional learning pipelines. This approach is designed to improve \textbf{robustness}, \textbf{interpretability}, and \textbf{domain generalization}, addressing critical limitations commonly encountered in medical deployments where data heterogeneity, distribution shifts, and limited supervision can degrade model performance.
\begin{figure*}[ht]
    \centering
    \includegraphics[width=0.8\textwidth]{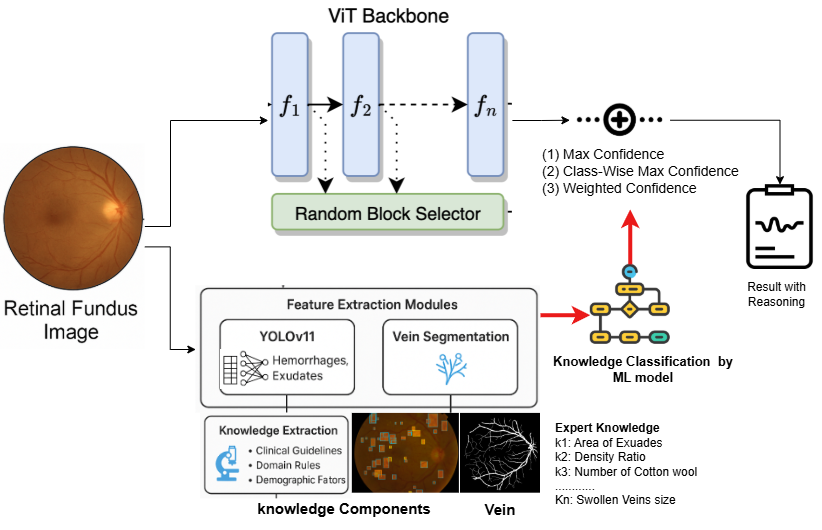}
    \caption{\small Overview of the proposed knowledge-guided DR classification framework, illustrating the integration of symbolic clinical rules and deep learning features \cite{kauppi2019}.}
    \label{fig:methodology}
\end{figure*}

\subsection{Knowledge-Guided Augmentation of Deep Models}

Traditional deep learning models typically learn a predictive mapping $f_{\text{DL}}: \mathcal{X} \rightarrow \mathcal{Y}$, where $\mathcal{X}$ denotes input modalities (e.g., retinal images) and $\mathcal{Y}$ represents target disease labels. This approach inherently lacks structured medical inductive biases, potentially limiting clinical applicability. To overcome this limitation, we propose a \textit{dual-branch architecture}, integrating structured knowledge representation $\mathcal{K}$ into deep learning-based image analysis.

We formalize $\mathcal{K}$ as a set of diagnostic rules $\{r_1, r_2, \dots, r_n\}$, each reflecting expert-validated correlations between observable clinical features and disease states. These rules incorporate visual biomarkers such as (e.g., exudates, hemorrhages, vascular patterns) and demographic parameters (e.g., patient age, glycemic status). For practical implementation, we develop corresponding \textit{feature extractors} $C = \{c_1, c_2, \dots, c_k\}$, instantiated via object detection models (YOLOv11), segmentation architectures, and logical rule functions.

Each extractor $c_i$ outputs a quantitative feature $f_i \in \mathbb{R}$, aggregated into a structured vector:

$$
F^* = \{f_1, f_2, \dots, f_k\}.
$$
This structured vector encodes clinical attributes such as presence, severity, and spatial distribution of significant retinal lesions, facilitating symbolic reasoning aligned closely with clinical diagnostic criteria.

A parallel \textit{knowledge-driven classifier} $f_{\text{KD}}: F^* \rightarrow \mathcal{Y}$ is trained alongside the deep learning model $f_{\text{DL}}$. The final prediction can then be determined through different fusion strategies. In the simplest case, a \textbf{selective fusion} rule is applied:

\[
y_{\text{final}} = 
\begin{cases}
y_{\text{DL}}, & \text{if } s_{\text{DL}} \geq s_{\text{KD}}, \\
y_{\text{KD}}, & \text{otherwise},
\end{cases}
\]

\noindent where $s_{\text{DL}}$ and $s_{\text{KD}}$ denote the maximum confidence scores from the deep and symbolic classifiers, respectively. This strategy enhances robustness by leveraging symbolic reasoning when the deep model predictions exhibit uncertainty, particularly valuable in handling out-of-distribution scenarios. Beyond this, we experimented with three additional fusion techniques:

1. \textbf{Max Confidence Fusion}: both the neural (ViT) and symbolic classifiers output calibrated probabilities via softmax normalization. The class with the globally highest confidence is selected, irrespective of source.  

2. \textbf{Class-wise Max Fusion}: normalized per-class confidence scores are compared across models, and the prediction is made according to the higher class-specific confidence.  

3. \textbf{Weighted Fusion}: empirically tuned weights $(\alpha_{DL}, \alpha_{KL})$ are applied to balance neural and symbolic predictions. Formally,
\[
y_{\text{final}} = \arg\max_{c \in \mathcal{C}} 
\left( \alpha_{DL} \cdot s_{DL}(c) + \alpha_{KL} \cdot s_{KL}(c) \right),
\]
where $s_{DL}(c)$ and $s_{KL}(c)$ are the softmax confidence scores assigned by the deep and symbolic classifiers, respectively, for class $c$, and $\mathcal{C}$ is the set of all DR severity classes.

Together, these strategies allow us to assess the trade-off between model confidence, robustness, and the influence of symbolic knowledge on final decision-making.

\subsection{ Diabetic Retinopathy Classification}

We evaluated the proposed KG-DG framework on the task of diabetic retinopathy (DR) classification using retinal fundus images—well-suited for knowledge-guided learning due to the presence of clearly defined visual pathologies such as microaneurysms, hemorrhages, exudates, and neovascularization. Domain-specific diagnostic rules were curated from ophthalmological guidelines (see Table~\ref{tab:dr_symptoms}) and operationalized via automated feature extraction pipelines built using two open-source, modular tools: YOLOv11 and a retinal vessel segmentation model.

For lesion-level localization, we employed the YOLOv11 object detection model, a state-of-the-art one-stage detector known for its efficiency and precision in dense object environments. YOLOv11 extends the YOLOv5/YOLOv7 series with advanced improvements including CSPDarkNet-based backbones, decoupled heads, and dynamic label assignment (DLA), achieving superior mean average precision (mAP) with real-time inference capabilities \cite{wang2023yolov11}. We fine-tuned YOLOv11 to detect clinically relevant lesions such as hemorrhages, hard exudates, and cotton wool spots. Bounding boxes produced by the model were post-processed and validated using Intersection over Union (IoU) scores against expert-labeled fundus images, ensuring medical fidelity.

In parallel, we integrated a vein segmentation module to extract morphological vessel features. This module, adapted from the open-source DRIVE and CHASE-DB1 datasets, uses a modified U-Net architecture with spatial attention layers to segment retinal vessels with high sensitivity. From these segmented maps, we extracted quantitative features including vessel tortuosity, branching angles, and average caliber—biomarkers strongly associated with DR progression. 

This structured knowledge vector was passed into a parallel symbolic classifier trained independently from the deep model, enabling our system to rely on rule-driven inference when the deep model exhibits uncertainty. Various machine learning models, including Logistic Regression, Random Forest, Support Vector Machines (SVM), Gradient Boosting, and K-Nearest Neighbors, were evaluated for knowledge-based classification on the feature set ($F$). Among these, Gradient Boosting demonstrated the best classification performance. Both YOLOv11 and the vein segmentation module functioned solely as independent auxiliary components to extract biomarkers from images, facilitating symbolic reasoning. The biomarkers were annotated by expert medical annotators on approximately 500 images, with random samples validated by respective domain experts. These annotations were subsequently used to fine-tune the YOLOv11 and vein segmentation modules, which then act as knowledge extractors within the pipeline. This accurate integration of clinical knowledge enhances model robustness, promotes domain invariance, and provides a solid foundation for understanding domain shifts through distributional alignment.

The classification results from the knowledge-based machine learning model and the ViT model are integrated using three main methods as shown in Figure 2: (1) selecting the maximum confidence score across all predictions, (2) computing the class-wise maximum confidence, and (3) applying a weighted confidence scheme. The outcomes of these three integration strategies are evaluated to assess the overall performance of the final framework.
%\vspace*{-1.5em} % remove 1 line of extra space
\subsection{Backbone Architectures and Training Strategy}
\vspace*{-0.3em} 
For the image-based analysis, we employed advanced Vision Transformer (ViT) architectures. 
The DeiT-small architecture, comprising approximately 22M parameters, was used without distillation \cite{touvron2021deit}. 
The CvT-13 model, with 20M parameters, integrates convolutional layers with transformer blocks to enhance spatial feature learning \cite{wu2021cvt}. 
Additionally, we utilized T2T-ViT-14, featuring progressive tokenization and encompassing 21.5M parameters \cite{yuan2021t2tvit}. 

All ViT models were initialized with ImageNet-pretrained weights, and during training, encoder parameters remained fixed to prevent overfitting. 
Only the classification heads underwent optimization using class-weighted cross-entropy loss. 
Training adhered to DomainBed protocols, employing resizing to 224×224, random cropping, horizontal flipping, color jitter, and grayscale augmentation. 
AdamW optimizer was utilized with a learning rate of $5 \times 10^{-5}$, and early stopping was implemented after 10 epochs without performance improvement.

\subsection{Evaluation Protocol and Results}

Initially, KG-DG is evaluated on the Aptos Dataset (60\% training, 20\% cross-validation, and 20\% testing), achieving superior performance, exceeding a ViT benchmark by 6\% (84.65\% vs. 78.40\%) and significantly outperforming existing baselines.  We conducted extensive evaluations in both multi-source and single-source domain generalization settings using publicly available DR datasets: APTOS \cite{kauppi2019}, EyePACS \cite{eyepacs2015}, MESSIDOR, and MESSIDOR2 \cite{messidor2014}. Each dataset constituted a distinct domain. In multi-source experiments, we trained models on three datasets while testing on the fourth. In single-source setups, we trained on a single dataset and evaluated on the remaining domains.

Our knowledge-guided framework consistently demonstrated superior performance, achieving a +2.1\% average accuracy improvement in multi-source domain generalization and a notable +4.2\% increase in single-source domain generalization scenarios, particularly impactful on imbalanced data distributions (see detailed results in Table 6).

The structured knowledge-driven classifier notably improved generalization by encapsulating domain-invariant medical reasoning, whereas the deep learning branch effectively modeled intricate visual patterns, validating the effectiveness of integrating clinical expertise within deep learning frameworks.

\textit{Note. Unless otherwise stated, in all tables the best-performing value within each column is highlighted in \textbf{bold}.}

\section{Experiments}
\subsection{Single Domain Generalization Results}
In the SDG setting, models were trained on one dataset and evaluated on the remaining three to simulate clinical deployment in unseen environments. Our method was evaluated against DRGen, ERM-ViT, SD-ViT, and SPSD-ViT using  APTOS \cite{kauppi2019}, EyePACS \cite{eyepacs2015}, Messidor-1 and Messidor-2. \cite{messidor2014} as source domains respectively. As shown in Tables 2-5, our method consistently outperformed existing baselines in three out of four training configurations.

For instance, when trained on APTOS, the Non-Weighted DL+KL fusion achieved the highest average accuracy (59.9\%), outperforming all transformer baselines and showing superior generalization to diverse domains like MESSIDOR2. Similarly, when trained on MESSIDOR2, the Weighted DL+KL fusion delivered a performance of 65.5\%, highlighting robustness against shifts in both demographic and imaging characteristics. These results validate that symbolic knowledge integration enables effective generalization from a single domain, crucial for low-resource clinical settings.
\vspace{1\baselineskip}
\begin{table}[ht]
\centering
\scriptsize
\captionsetup{font=footnotesize} 
\caption{SDG Trained on APTOS - Cross-domain Accuracy (\%)}
\begin{tabular}{lcccc}
\hline
\textbf{Method} & \textbf{Eyepacs} & \textbf{Messidor} & \textbf{Messidor2} & \textbf{Average} \\
\hline
DRGen & 67.5$\pm$1.8 & 46.7$\pm$0.1 & \textbf{61.0}$\pm$0.1 & 58.4$\pm$0.57 \\
ERM-ViT & 67.8$\pm$1.4 & 45.5$\pm$0.2 & 58.8$\pm$0.4 & 57.3$\pm$0.76 \\
SD-ViT & 72.0$\pm$0.8 & 45.4$\pm$0.1 & 58.5$\pm$0.2 & 58.6$\pm$0.22 \\
SPSD-ViT & 71.4$\pm$0.8 & 45.6$\pm$0.1 & 58.8$\pm$0.2 & 58.6$\pm$0.42 \\
\hline
VIT (DL) & 66.6$\pm$0.4 & 46.4$\pm$0.3 & 48.9$\pm$0.2 & 53.9$\pm$0.5 \\
Knowledge (KL) & 66.4$\pm$0.8 &  \textbf{49.6}$\pm$0.2 & 53.9$\pm$0.7 & 56.6$\pm$0.3 \\
Non Weighted (DL + KL) & \textbf{72.8}$\pm$0.5 & 50.6$\pm$0.4 & 54.3$\pm$0.4 & \textbf{59.9}$\pm$0.2 \\
Weighted (DL + KL) & 67.4$\pm$0.3 & 49.6$\pm$0.3 & 53.9$\pm$0.6 & 57.0$\pm$0.2 \\
\hline
\end{tabular}
\end{table}

\begin{table}[ht]
\centering
\scriptsize
\captionsetup{font=footnotesize} 
\caption{SDG Trained on MESSIDOR - Cross-domain Accuracy (\%)}
\begin{tabular}{lcccc}
\hline
\textbf{Method} & \textbf{Aptos} & \textbf{Eyepacs} & \textbf{Messidor2} & \textbf{Average} \\
\hline
DRGen & 41.7$\pm$4.3 & 43.1$\pm$7.9 & 44.8$\pm$0.9 & 43.2$\pm$0.65 \\
ERM-ViT & 45.3$\pm$1.3 & 52.4$\pm$3.2 & 58.2$\pm$3.2 & 51.9$\pm$0.71 \\
SD-ViT & 44.3$\pm$0.9 & 53.2$\pm$1.6 & 57.8$\pm$2.4 & 51.7$\pm$0.35 \\
SPSD-ViT & 48.3$\pm$1.1 & 57.4$\pm$2.1 & 62.2$\pm$1.6 & 55.9$\pm$0.88 \\
\hline
VIT (DL) & 49.8$\pm$0.4 & \textbf{62.1}$\pm$0.3 & 59.1$\pm$0.3 & 57.0$\pm$0.5 \\
Knowledge (KL) & \textbf{74.0}$\pm$0.5 & 63.6$\pm$0.4 & \textbf{63.8}$\pm$0.3 & \textbf{67.1}$\pm$0.2 \\
Non Weighted (DL + KL) & 52.7$\pm$0.7 & 63.4$\pm$0.4 & 61.4$\pm$0.5 & 59.2$\pm$0.4 \\
Weighted (DL + KL) & 74.1$\pm$0.5 & 63.3$\pm$0.2 & 63.8$\pm$0.6 & 67.1$\pm$0.7 \\
\hline
\end{tabular}
\end{table}

\begin{table}[H]
\centering
\scriptsize
\captionsetup{font=footnotesize} 
\caption{SDG Trained on MESSIDOR2 - Cross-domain Accuracy (\%)}
\begin{tabular}{lcccc}
\hline
\textbf{Method} & \textbf{Aptos} & \textbf{Eyepacs} & \textbf{Messidor} & \textbf{Average} \\
\hline
DRGen & 40.9$\pm$3.9 & 69.3$\pm$1.0 & 61.3$\pm$0.8 & 57.7$\pm$0.67 \\
ERM-ViT & 47.9$\pm$2.1 & 67.4$\pm$0.9 & 59.6$\pm$3.9 & 58.3$\pm$0.33 \\
SD-ViT & 51.8$\pm$0.9 & 68.7$\pm$0.6 & \textbf{62.0}$\pm$1.7 & 60.8$\pm$0.58 \\
SPSD-ViT & 52.8$\pm$2.0 & \textbf{72.5}$\pm$0.3 & 61.0$\pm$0.8 & 62.1$\pm$0.85 \\
\hline
VIT (DL) & 29.2$\pm$0.4 & 44.7$\pm$0.5 & 49.4$\pm$0.7 & 41.1$\pm$0.7 \\
Knowledge (KL) & \textbf{69.1}$\pm$0.3 & 71.1$\pm$0.4 & 55.3$\pm$0.9 & 65.2$\pm$0.5 \\
Non Weighted (DL + KL) & 63.6$\pm$0.6 & 71.1$\pm$0.8 & 56.4$\pm$0.2 & 63.7$\pm$0.6 \\
Weighted (DL + KL) & 69.5$\pm$0.4 & 71.0$\pm$0.2 & 55.9$\pm$0.6 & \textbf{65.5}$\pm$0.3 \\
\hline
\end{tabular}
\end{table}

\begin{table}[H]
\centering
\scriptsize
\captionsetup{font=footnotesize} 
\caption{SDG Trained on EYEPACS - Cross-domain Accuracy (\%)}
\begin{tabular}{lcccc}
\hline
\textbf{Method} & \textbf{Aptos} & \textbf{Messidor} & \textbf{Messidor2} & \textbf{Average} \\
\hline
DRGen & 61.3$\pm$1.9 & 54.6$\pm$1.5 & 65.4$\pm$0.1 & 60.4$\pm$0.25 \\
ERM-ViT & 69.1$\pm$1.4 & 50.4$\pm$0.3 & 62.8$\pm$0.2 & 60.8$\pm$0.58 \\
SD-ViT & 69.3$\pm$0.3 & 50.0$\pm$0.5 & 62.9$\pm$0.2 & 60.7$\pm$0.41 \\
SPSD-ViT & \textbf{75.1}$\pm$0.5 & 50.5$\pm$0.8 & 62.2$\pm$0.4 & \textbf{62.5}$\pm$0.62 \\
\hline
VIT (DL) & 49.7$\pm$0.9 & 52.9$\pm$0.2 & 49.1$\pm$0.9 & 50.6$\pm$0.4 \\
Knowledge (KL) & 60.2$\pm$0.2 & \textbf{53.7}$\pm$0.6 & 66.5$\pm$0.4 & 60.13$\pm$0.5 \\
Non Weighted (DL + KL) & 63.9$\pm$0.2 & 53.8$\pm$0.3 & \textbf{67.2}$\pm$0.6 & 61.7$\pm$0.4 \\
Weighted (DL + KL) & 60.2$\pm$0.3 & 48.7$\pm$0.2 & 66.4$\pm$0.7 & 58.4$\pm$0.9 \\
\hline
\end{tabular}
\end{table}
%%%%%%%%%%%%%%%%%%%%%%%%%%%%%%%%%%%%%%%%%%%%%%%%%%%%%%%%%%%%%%%%%%%%%%%%

\subsection{Multi Domain Generalization Results}
In the MDG setting, we trained our model on three datasets and evaluated on the unseen fourth, as per the DomainBed protocol. Results in Table~\ref{tab:results} show that our KG-DG model using Clip-ViT (ViT+KL) and symbolic classifiers significantly improved generalization compared to popular convolutional and transformer-based DG methods, including ERM, IRM, Fishr, and SD-ViT.
Notably, the knowledge-guided symbolic model (KL only) achieved the best average accuracy (63.67\%), while SPSD-ViT and ERM-ViT with strong augmentations reached 65.5\%. Despite having fewer parameters, our model’s performance indicates effective utilization of symbolic lesion features and their generalization power across domain shifts. In particular, the KL model exceeded both standard ViT and ResNet baselines across most target domains, demonstrating the critical role of encoded clinical knowledge in cross-domain settings.

\renewcommand{\arraystretch}{1.2}
\setlength{\tabcolsep}{5pt}
\begin{table*}[h]
    \centering
     \caption{Performance comparison of different methods and backbones across diabetic retinopathy datasets (Accuracy \%).}
    \footnotesize
    \begin{tabular}{l l c c c c c}
        \toprule
        \textbf{Method} & \textbf{Backbone} (\#Param) & \textbf{Aptos} & \textbf{Eyepacs} & \textbf{Messidor} & \textbf{Messidor 2} & \textbf{Avg.} \\
        \midrule
        ERM \cite{vapnik1999nature} & ResNet50\textsubscript{(23.5M)} & 47.6$\pm$1.7 & 71.3$\pm$0.3 & 63.0$\pm$0.4 & 69.0$\pm$1.5 & 62.7 \\
        IRM \cite{arjovsky2019invariant} & ResNet50 & 52.1$\pm$1.7 & 73.2$\pm$0.3 & 51.3$\pm$3.8 & 57.2$\pm$1.7 & 58.4 \\
        ARM \cite{zhang2021adaptive} & ResNet50 & 45.6$\pm$1.5 & 71.7$\pm$0.5 & 62.4$\pm$1.0 & 60.0$\pm$3.4 & 59.9 \\
        Fish \cite{shi2021gradient} & ResNet50 & 44.6$\pm$2.2 & 72.7$\pm$0.7 & 62.1$\pm$0.7 & 66.4$\pm$1.7 & 61.4 \\
        Fishr \cite{rame2022fishr} & ResNet50 & 47.0$\pm$1.8 & 71.9$\pm$0.6 & 63.3$\pm$0.5 & 66.4$\pm$0.2 & 62.2 \\
        GroupDRO \cite{sagawa2020distributionally} & ResNet50 & 44.9$\pm$3.8 & 72.0$\pm$0.3 & 63.1$\pm$0.9 & 67.8$\pm$1.9 & 62.0 \\
        MLDG \cite{li2018learning} & ResNet50 & 44.1$\pm$1.6 & 72.7$\pm$0.6 & 62.7$\pm$0.6 & 64.4$\pm$0.4 & 61.0 \\
        Mixup \cite{yan2020improve} & ResNet50 & 47.3$\pm$1.7 & 72.0$\pm$0.3 & 59.8$\pm$2.8 & 65.8$\pm$1.4 & 61.2 \\
        Coral \cite{sun2016deep} & ResNet50 & 49.8$\pm$1.0 & 71.7$\pm$0.9 & 58.6$\pm$2.8 & 68.2$\pm$0.6 & 62.1 \\
        MMD \cite{li2018domain} & ResNet50 & 49.3$\pm$1.0 & 69.3$\pm$1.1 & 64.1$\pm$4.8 & 69.6$\pm$0.6 & 63.1 \\
        DANN \cite{ganin2016domain} & ResNet50 & \textbf{54.4$\pm$0.8} & 72.9$\pm$1.4 & 57.0$\pm$1.1 & 58.6$\pm$1.7 & 60.7 \\
        CDANN \cite{li2018conditional} & ResNet50 & 48.1$\pm$0.7 & 73.1$\pm$0.3 & 55.8$\pm$1.8 & 61.2$\pm$1.3 & 59.5 \\
        \midrule
        ERM-ViT \cite{vapnik1999nature} & DeiT-Small\textsubscript{(22M)} & 48.5$\pm$0.9 & 70.7$\pm$1.7 & 62.7$\pm$1.6 & 69.5$\pm$2.5 & 62.9 \\
        ERM-ViT \cite{vapnik1999nature} & T2T-14\textsubscript{(21.5M)} & 54.0$\pm$3.0 & 73.2$\pm$0.4 & 60.8$\pm$1.7 & 72.0$\pm$0.2 & 62.5 \\
        ERM-ViT \cite{vapnik1999nature} & CvT-13\textsubscript{(20M)} & 51.3$\pm$1.7 & 73.3$\pm$0.2 & 64.8$\pm$0.6 & 72.4$\pm$0.6 & 65.5 \\
        SD-ViT \cite{sultana2022self} & DeiT-Small\textsubscript{(22M)} & 48.2$\pm$2.5 & 69.6$\pm$1.5 & 63.9$\pm$1.3 & 65.0$\pm$1.7 & 61.8 \\
        SD-ViT \cite{sultana2022self} & T2T-14\textsubscript{(21.5M)} & 46.5$\pm$0.8 & 71.1$\pm$0.7 & 63.9$\pm$0.9 & 71.4$\pm$0.2 & 63.2 \\
        SPSD-ViT \cite{jayanga2023generalizing} & DeiT-Small\textsubscript{(22M)} & 51.6$\pm$1.1 & 73.3$\pm$0.4 & 64.0$\pm$1.4 & \textbf{72.9$\pm$0.1} & 65.5 \\
        SPSD-ViT \cite{jayanga2023generalizing} & T2T-14\textsubscript{(21.5M)} & 50.0$\pm$2.8 & \textbf{73.6$\pm$0.3} & \textbf{65.2$\pm$0.3} & \textbf{73.3$\pm$0.2} & \textbf{65.5} \\
        SPSD-ViT \cite{jayanga2023generalizing} & CvT-13\textsubscript{(20M)} & \textbf{51.7$\pm$1.2} & 73.3$\pm$0.2 & 64.8$\pm$0.6 & 72.4$\pm$0.6 & 65.5 \\
        \midrule
        \textbf{ViT (Ours)} & Vit \textsubscript{(22M)} & 50.1$\pm$1.7 & 69.4$\pm$0.3 & 58.13$\pm$3.8 & 67.1$\pm$1.7 & 61.18 \\
        \textbf{ViT +KL (Ours)} & Vit\textsubscript{(21.5M)} & 53.1$\pm$1.7 & 72.2$\pm$0.3 & 51.3$\pm$3.8 & 56.2$\pm$1.7 & 58.4 \\ 
        \textbf{KL (Ours)} & Knowledge\textsubscript{(20M)} & \textbf{60.70$\pm$1.2} & 68.45$\pm$0.2 & 58.67$\pm$0.6 & 67.66$\pm$0.6 & 63.67 \\
        \bottomrule
    \end{tabular}
   
    \label{tab:results}
\end{table*}

\section{Evaluation}

\subsection{Benchmark Setup}

To rigorously evaluate the generalization capability of the proposed KG-DG framework, we conducted experiments on four publicly available diabetic retinopathy (DR) fundus image datasets: APTOS \cite{kauppi2019}, EyePACS, Messidor-1, and Messidor-2. Each dataset represents a distinct clinical domain, differing significantly in patient demographics, imaging devices, and image acquisition protocols. Following the DomainBed benchmark protocol established by Gulrajani et al. \cite{gulrajani2021}, we implemented two experimental scenarios: Single-Domain Generalization (SDG), wherein the model is trained on a single domain and evaluated on the remaining three domains, and Multi-Domain Generalization (MDG), where training is performed on three domains with evaluation conducted on a separate unseen domain.

For preprocessing, all images were uniformly resized to \(224 \times 224\) pixels and subjected to data augmentations including center cropping, horizontal flipping, color jittering, and grayscale conversion to mimic realistic variability and prevent dataset-specific biases. To ensure a robust and unbiased evaluation, early stopping was applied based on validation accuracy computed on the training domain(s).

\subsection{Baseline Models}

We evaluated our KG-DG framework against several competitive baseline methods representative of both convolutional neural network (CNN)-based and transformer-based domain generalization strategies. For convolutional architectures, we included Empirical Risk Minimization (ERM) with ResNet-50 \cite{he2016resnet}, a strong baseline under fair evaluation standards \cite{gulrajani2021}. Additionally, we compared against Invariant Risk Minimization (IRM) \cite{arjovsky2019invariant}, Group Distributionally Robust Optimization (GroupDRO) \cite{sagawa2020distributionally}, Fishr \cite{rame2022fishr}, and Adaptive Risk Minimization (ARM) \cite{zhang2021adaptive}, each employing distinct strategies to enforce robustness and domain invariance.

Transformer-based models considered included ERM-ViT with DeiT-Small \cite{touvron2021}, CvT-13 \cite{wu2021cvt}, and T2T-ViT \cite{yuan2021t2t}. We further included state-of-the-art transformer-based domain generalization models, SD-ViT \cite{sultana2022self} and SPSD-ViT \cite{jayanga2023generalizing}, which utilize semantic alignment and pseudo-labeling to enhance robustness. Lastly, we compared against DRGen \cite{atwany2022drgen}, a DR-specific DG method leveraging adversarial and contrastive learning.

The evaluation of our framework and comparative methods was performed using multiple metrics designed to comprehensively assess the models’ performance under domain shift. Cross-domain accuracy was employed as the primary metric to gauge generalization effectiveness on unseen datasets. To address inherent class imbalance common in diabetic retinopathy classification tasks, we reported the Macro F1-score, which provides a balanced measure across all DR severity classes. Additionally, we calculated the Area Under the Receiver Operating Characteristic Curve (AUC-ROC), offering insights into sensitivity-specificity trade-offs critical in medical diagnostics.

To quantify distributional alignment across domains, we employed KL divergence between domain-specific embeddings. Lastly, the quality and reliability of our symbolic lesion detection modules were assessed through Intersection-over-Union (IoU) scores against expert annotations, ensuring clinical relevance and interpretability of the symbolic knowledge incorporated into our framework.
\subsection{Ablation Study}

\textbf{Ablation Study I: APTOS-Trained Domain Generalization} To understand the individual and combined contributions of neural and symbolic components in our framework, we conducted a focused ablation study using the APTOS dataset as the source domain. Table~\ref{tab:aptos_ablation} reports the accuracy performance on three unseen target domains—EyePACS, Messidor-1 and Messidor-2 when models were trained solely on APTOS.

The neural-only baseline using Vision Transformer (ViT) achieves a modest average accuracy of 53.9\%, indicating limited generalization under domain shift. The symbolic-only model, based on knowledge-driven lesion features (KL), improves the average accuracy to 56.6\%, highlighting the value of structured clinical priors. The best performance is observed when combining both neural and symbolic reasoning. In particular, the non-weighted fusion approach yields the highest average accuracy of 59.9\%, outperforming both standalone models. This result demonstrates the strength of the proposed neuro-symbolic integration in improving robustness and domain generalization in diabetic retinopathy classification.
\vspace{0.5em}

\begin{table}[ht]
\centering
\caption{Ablation study comparing neural-only (ViT), symbolic-only (KL), and fused (DL+KL) models, trained on the APTOS dataset and evaluated on unseen domains. Neuro-symbolic fusion achieves the highest average generalization accuracy.}
\begin{tabular}{|l|c|c|c|c|}
\hline
\textbf{Setting} & \textbf{Eyepacs} & \textbf{Messidor} & \textbf{Messidor2}   \\
\hline
\textbf{Neural Only (ViT)} & 66.6 & 46.4 & 48.9  \\
\textbf{Symbolic Only (KL)} & 66.4 & 49.6 & 53.9  \\
\textbf{Neural + Symbolic (Non-Weighted)} & \textbf{72.8} & \textbf{50.6} & \textbf{54.3}  \\
\textbf{Neural + Symbolic (Weighted)} & 67.4 & 49.6 & 53.9   \\
\hline
\end{tabular}
\label{tab:aptos_ablation}
\end{table}

\begin{table*}[h]
\centering
\caption{
\textbf{Ablation Study on Symbolic Lesion Biomarkers with and without Retinal Vein Features.}
The first section evaluates performance with lesion biomarkers alone \textit{exudates, hard hemorrhages, soft hemorrhages, and cotton wool spots} on the APTOS dataset; the second includes additional retinal vein morphology features (e.g., tortuosity, caliber, branching angles).  
}
\vspace{0.5em}
\resizebox{\textwidth}{!}{
\begin{tabular}{l|c|c|c|c|c|c|c|c}
\toprule
\textbf{Model} & \textbf{Feature Set} & \textbf{Accuracy} & \textbf{F1-Score} & \textbf{Precision} & \textbf{Recall} & \textbf{Exudate Score} & \textbf{Hemorrhage Score} & \textbf{AUC} \\
\midrule
Logistic Regression & Lesions Only & 0.7732 & 0.7322 & 0.59 & 0.49 & 0.77 & 0.75 & 0.74 \\
Random Forest & Lesions Only & 0.8169 & 0.8115 & 0.82 & 0.80 & 0.80 & 0.78 & 0.81 \\

SVM                  & Lesions Only & 0.7814 & 0.7432 & 0.59 & 0.50 & 0.77 & 0.75 & 0.76 \\
Gradient Boosting    & Lesions Only & \textbf{0.8465} & \textbf{0.8412} & \textbf{0.82} & \textbf{0.76} & \textbf{0.83} & \textbf{0.80} & \textbf{0.84}
\\
K-Nearest Neighbors  & Lesions Only & 0.7814 & 0.7896 & 0.63 & 0.56 & 0.78 & 0.76 & 0.77 \\
\midrule
Logistic Regression  & Lesions + Vein & 0.6424 & 0.6019 & 0.25 & 0.33 & 0.55 & 0.58 & 0.58 \\
Random Forest        & Lesions + Vein & 0.7384 & 0.7038 & 0.55 & 0.47 & 0.71 & 0.71 & 0.70 \\
SVM                  & Lesions + Vein & 0.6556 & 0.6083 & 0.26 & 0.34 & 0.56 & 0.59 & 0.58 \\
Gradient Boosting    & Lesions + Vein & 0.7252 & 0.7389 & 0.51 & 0.44 & 0.70 & 0.70 & 0.69 \\
K-Nearest Neighbors  & Lesions + Vein & 0.6987 & 0.6369 & 0.43 & 0.44 & 0.65 & 0.67 & 0.66 \\
\bottomrule
\end{tabular}
}
\label{tab:ablation}
\end{table*}

\textbf{Ablation Study II: Performance of Symbolic Lesion Biomarkers with and without Retinal Vein Features.}
This experiment evaluates the discriminative capacity of structured symbolic features extracted from retinal images, focusing on four clinically validated lesion types: \textit{exudates, hard hemorrhages, soft hemorrhages, and cotton wool spots}. The first group of results in Table~\ref{tab:ablation} includes only lesion-based features, while the second incorporates additional vascular information derived from retinal vein morphology—such as tortuosity, caliber, and branching angles. 

Across all classifiers, models trained solely on lesion features consistently outperform those that include both lesions and vein information. Gradient Boosting achieves the highest accuracy (84.65\%) and macro F1-score (84.12\%), confirming the strong discriminative value of lesion-level biomarkers. In contrast, the addition of vein-based features leads to performance degradation, indicating that vessel morphology introduces domain-sensitive variability that hampers generalization. 

Accordingly, our main KG-DG framework prioritizes lesion biomarkers as the most reliable symbolic inputs, while vein features are treated as optional. This design choice also strengthens interpretability: lesion counts and distributions directly align with established clinical diagnostic protocols, whereas vessel morphology requires context-specific calibration and exhibits less transferability across domains.
\subsection{Discussion and Limitations}
The KG-DG framework achieves consistent generalization across unseen domains by combining symbolic clinical knowledge with deep visual features, though several limitations remain. Its reliance on accurate lesion-level annotations and pre-trained modules like YOLOv11 and retinal vein segmentation introduces dependency on expert-verified data, which may not be available for other medical imaging tasks. The symbolic classifier may miss complex visual cues, and fusion performance varies with confidence-weighting strategies, highlighting a need for more adaptive mechanisms. Across SDG and MDG tasks, maximum cross-domain improvement reaches 5.2\% (Messidor2 $\rightarrow$ APTOS), with average gains around 2--3\%, indicating steady but not uniform dominance. Feature importance analysis shows lesion features such as ``exudates count'' and ``hemorrhage density'' align with clinical practice, supporting human-aligned decision-making. Future work could explore dynamic neuro-symbolic reasoning, integrate temporal clinical data, and extend KG-DG to other modalities like OCT or histopathology.

\section{Conclusions}

This paper introduces KG-DG, an improved knowledge-guided domain generalization framework specifically tailored for medical imaging applications, as exemplified by diabetic retinopathy classification. KG-DG integrates symbolic clinical reasoning and deep visual representations through a confidence-weighted fusion approach, significantly enhancing robustness and interpretability. Comprehensive experimental results on four diverse DR datasets demonstrated that KG-DG consistently achieved superior performance compared to strong baselines of domain generalization methods, achieving notable improvements in both single-source and multi-source generalization settings, with gains of up to 5.2\% accuracy in cross-domain accuracy.

Our findings underscore the importance of embedding structured clinical knowledge within deep learning models, thereby significantly improving generalization and trustworthiness in clinical settings. Future directions include adapting the KG-DG framework to additional medical imaging modalities, such as optical coherence tomography and histopathology, and further integrating dynamic symbolic reasoning via neuro-symbolic architectures, enhancing real-time decision support capabilities in medical AI deployments. \textbf{Insights:} Our observations indicate that the integration of symbolic clinical knowledge into traditional architectures—whether Vision Transformers (ViTs) or domain-specific models such as DeepXSOZ \cite{shama2023deepsoz}—consistently leads to significant improvements in classification accuracy. Furthermore, this knowledge imputation enhances both domain generalization and the explainability of model behavior, addressing critical challenges in clinical deployment.

\section*{Acknowledgments}

This work was partly funded by NSF (FDT-Biotech 2436801), and the Helmsley Charitable Trust (2-SRA-2017-503-M-B).

%%%%%%%%%%%%%%%%%%%%%%%%%%%%%%%%%%%%%%%%%%%%%%%%%%%%%%%%%%%%%%%%%%%%%%%%

%%% Use this environment to include acknowledgements (optional).
%%% This will be omitted in doubleblind mode.

%%%%%%%%%%%%%%%%%%%%%%%%%%%%%%%%%%%%%%%%%%%%%%%%%%%%%%%%%%%%%%%%%%%%%%%%

%%% Use this command to include your bibliography file.

\bibliography{ecai-sample-and-instructions}

\end{document}